\newcommand{\non}{\nonumber}
\newtheorem{theorem}{Theorem}

\def\T{\tiny\mbox{\rm T}}

\def\bfH{\bm H}

\documentclass[fleqn,letterpaper, 10 pt, conference]{ieeeconf}  

\IEEEoverridecommandlockouts                              
\usepackage{cite}
\usepackage{amsmath,amssymb,amsfonts}
\usepackage{cases}
\usepackage{algorithmic}
\usepackage{graphicx}
\usepackage{graphicx}
\usepackage{subcaption}
\usepackage{textcomp}
\usepackage{bm}
\usepackage{theorem}
\usepackage{color}
 \usepackage{makecell}
 \usepackage{algorithm}
 \usepackage{multirow}
 \newtheorem{Definition}[theorem]{Definition}
\allowdisplaybreaks  

\title{\textbf{Identification of Non-causal Graphical Models}}
\author{Junyao You, Mattia Zorzi
\thanks{This work was supported by Chongqing Natural Science Foundation CSTB2023NSCQ-JQX0018, National Natural Science Foundation of China under Grant 61991414, Beijing Natural Science Foundation L221005, and the China Scholarship Council scholarship.}
\thanks{J. You is with the School of Automation, Beijing Institute of Technology,
Beijing 100081, China; {{\tt\small yjy804521297@163.com}}.}
\thanks{M. Zorzi is with the Department of Information Engineering, University of Padova, Via Gradenigo 6/B, 35131 Padova, Italy;  {{\tt\small zorzimat@dei.unipd.it}}.}
}

\begin{document}

\maketitle
\thispagestyle{empty}
\pagestyle{empty}

\begin{abstract}
The paper considers the problem to estimate non-causal graphical models whose edges encode smoothing relations among the variables.
We propose a new covariance extension problem and show that  the solution minimizing the transportation distance with respect to white noise process is a double-sided autoregressive non-causal graphical model.
Then, we generalize the paradigm to a class of graphical autoregressive moving-average models.
Finally, we test the performance of the proposed method through some
numerical experiments.
\end{abstract}

\section{Introduction}
\label{sec:introduction}
 Graphical models
 find their applications in many fields such as bioinformatics, image processing, finance and econometrics,
 see for instance \cite{Jordan2004-GM,Kock2015-Ec,Alfred2009-SI}.
 The latter provide convenient tools to visualize and infer the relations among the involved variables.
 Given their rich application scenarios, numerous research efforts have been undertaken to learn such models from data, \cite{TACZHU24,HOF1,BKRON,tugnait2024learning}.

One possible way to learn a dynamic graphical model is to consider a covariance extension problem \cite{RKL-16multidimensional,enqvist2004aconvex,RFP-09,FRT-11,zhu2018wellposed}, i.e.
the problem to find a power spectral density matching some covariance lags which are typically inferred from data.
This approach has been exploited to derive  paradigms  for the identification of autoregressive (AR) graphical models \cite{SONGSIRI_TOP_SEL_2010,ZORZI2019108516,MAANAN2017122,RECS,zorzi2019graphical}, autoregressive moving-average (ARMA) graphical models\cite{6365751,e20010076,YOU2022110319,Mattia2023ARMA,LINKPRED,You2022ARMA}, latent-variable graphical models\cite{LATENTG,ALPAGO_SL_REC,you2023sparse,You2023SLARMA,TAC19} and Kronecker graphical models \cite{KRON}.
All these paradigms consider graphical models whose edges encode conditional dependence relations among the variables.

An important class of dynamic directed networks is the one whose edges encode smoothing relations among the variables, \cite{M2012-WF,9600870}.
In such models, the absence of the directed edge from node $l$ to node $i$ means that
the variable corresponding to node $l$ is not needed for computing the smoothing estimate of the variable corresponding to node $i$.
To the best of our knowledge, there has been no work relating the covariance extension problem to this type of dynamic network.

In this paper we restrict our attention to symmetric directed non-causal graphical models whose edges appear twice, one in each direction, and encode smoothing relations. We show that it is possible to  consider a covariance extension problem for which a particular solution is a graphical model of this type.
More precisely, the latter is the one minimizing the transportation distance, \cite{Z-TRANSP}, with respect to white Gaussian noise.
Moreover, such solution is  a double-sided AR model.
We also extend this paradigm to a  class of ARMA models.

We warn the reader that the present paper only reports some preliminary result regarding the estimation of non-causal graphical models. In particular, all the proofs and most of the technical assumptions needed therein are omitted and will be published afterwards.
The rest of the paper is organized as follows:
we introduce the problem of estimating non-causal graphical models in Section \ref{sec:De_GM}.
In Section \ref{sec:TS} we propose the covariance extension problem based on the transportation distance, while in Section \ref{sec:TS_dual} we introduce the corresponding dual problem showing that the solution to the primal problem is a non-causal graphical model.
Section \ref{sec:TS_ARMA} extends the results for the ARMA case. Section \ref{sec:Simulation} contains some
numerical experiments to test the performance of the proposed
estimators. Finally, the conclusions are drawn in Section \ref{sec:Conclusions}.

\section{Non-causal graphical models}
\label{sec:De_GM}
Consider the discrete-time stationary  double-sided AR model process $y:=\{\, y(t),\; t\in\mathbb Z\}$ defined as:
\begin{eqnarray}
\label{non_causal_AR}
y(t)&&=H(z)y(t)+e(t),\\
\label{H_z}
H(z)&&=H_0+\frac{1}{2}\sum_{k=1}^{n}({H_k}z^{-k}+{H_k}^{\T}z^{k}),\non
\end{eqnarray}
where $y(t)\in\mathbb{R}^{m}$ is the output of the system,
$e(t)\in\mathbb{R}^{m}$ is Gaussian white noise with zero mean and spectral density $\Phi_e=I_m$,
and $H(z)$ is a polynomial matrix such that $H_k\in\mathbb{R}^{m\times m}$, with $k=0,\ldots,n$, $[H_k]_{ll}=0$, with $l=1,\ldots,m$,
and $H_0$ is symmetric. Thus, we have $H(z)=H(z^{-1})^{\top}$.
We assume that
$I_m-H(z)>0$ for any $z \in \mathbb{C}$ with $|z|=1$. In this way, process $y$ admits the following ``forward'' representation
$$ y(t)=[I_m-H(z)]^{-1}e(t)$$
and its spectral density is a function defined on the unit circle $\{e^{{\rm{j}} \theta}\ \rm{s.t.}\ \theta\in[-\pi,\pi]\}$ taking the form
\begin{eqnarray}
\label{Phi1}
\Phi(e^{{\rm{j}} \theta})=[I_m-H(e^{{\rm{j}} \theta})]^{-2}.
\end{eqnarray}
In the following the dependence upon $\theta$ will be
dropped if not needed, e.g. $\Phi$ instead of $\Phi(e^{{\rm{j}} \theta})$.

Let $y_l:=\{\, y_l(t),\; t\in\mathbb Z\}$, with $l=1\ldots m$, denote the $l$-th component of $y$. In what follows we assume that the support of $H(z)$ corresponds to a self-kin graph, then smoothing estimate $\hat{y}_l(t)$ of $y_l(t)$ given the other components of $y$ is defined as
\begin{eqnarray}
\hat{y}_l(t)=\sum_{l\neq i}[H(z)]_{li}{y}_i(t),\non
\end{eqnarray}
where $[X]_{li}$ represents the entry of row $l$ and column $i$ in matrix $X$.
It can be seen that if  $[H(z)]_{li}=0$,
we do not need the $i$-th component of $y$ in order to compute $\hat{y}_l(t)$. On the contrary, if $[H(z)]_{li}\neq 0$, then we need the $i$-th component of $y$ in order to compute $\hat{y}_l(t)$. Since $H(z)=H(z^{-1})^{\top}$, then
$[H(z)]_{li}=0$ if and only if $[H(z)]_{il}=0$. Accordingly,
we can attach a symmetric directed graph $\mathcal G(\mathcal V,\mathcal E)$ to Model (\ref{non_causal_AR}) with vertex set $\mathcal V=\{1,\ldots,m\}$  and edge set $\mathcal E\subseteq \mathcal V\times \mathcal V$ such that $(i,l)\in \mathcal E$ and $(l,i)\in \mathcal E$ if and only if $[H(z)]_{li}\neq 0$, i.e. the $i$-th component of $y$ is needed for computing the smoothing estimate of the $l$-th component and vice versa.

\begin{figure}[!hbt]
  \centering
  \includegraphics[width=0.2\hsize]{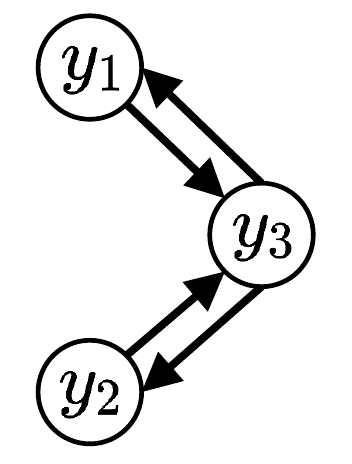}
  \caption{Graph corresponding to Model (\ref{non_causal_AR}) with $H(z)$ defined in (\ref{defH}). Every node represents a component of process $y$. We have: two edges between $y_1$ and $y_3$ because $[H(z)]_{13}\neq 0$; two edges between $y_2$ and $y_3$ because $[H(z)]_{23}\neq 0$.}
  \label{Fig_non_causal_AR_ex}
\end{figure}

\textbf{Example.} Consider Model (\ref{non_causal_AR}) with $m=3$ and
\begin{eqnarray}\label{defH}
H(z)
=\setlength\arraycolsep{2pt}
        \left[\begin{array}{ccc}
           0 & 0  &  [H(z)]_{13}  \\
             0    &0    &  [H(z)]_{23}\\
             {[H(z^{-1})]_{13}}    & [H(z^{-1})]_{23}   &  0       \end{array}\right]
\end{eqnarray}
with $[H(z)]_{13}$ and $[H(z)]_{23}$ different from the null function.
Then, the corresponding graph is depicted in Fig. \ref{Fig_non_causal_AR_ex}.

It is worth noting that $\hat y_l(t)$ is the non-causal Wiener filter of $y_l(t)$ given the other components of $y$:
$$ \hat y_l(t)=\mathbb E[ y_l(t)| \mathcal Y_{\mathcal V}],$$
where $\mathbb E[\cdot|\cdot]$ denotes the conditional expectation operator and
$$ \mathcal Y_{\mathcal I}:=\overline{\mathrm{span}}\{ y_i(t) \hbox{ s.t. } i\in \mathcal I, \; t\in\mathbb Z \}$$ is the closure of all finite linear combination of $y_i(t)$ with $i\in \mathcal I\subseteq \mathcal V$ and $t\in\mathbb Z$. Notice that $\mathcal Y_{\mathcal I}$ is a vector subspace of the Hilbert space of Gaussian random variables having finite second order moments, \cite{LP15}. Therefore,
 $[H(z)]_{li}\equiv 0$ if and only if the following conditions hold
\begin{align*}
\mathbb E[y_l(t)|\mathcal Y_{\mathcal V}]&=\mathbb E[y_l(t)|\mathcal Y_{\mathcal V\setminus\{i,l\}}],\\
\mathbb E[y_i(t)|\mathcal Y_{\mathcal V}]&=\mathbb E[y_i(t)|\mathcal Y_{\mathcal V\setminus\{i,l\}}],
\end{align*}
and we say that the $i$-th and $l$-th components of $y$ are non-causally Wiener independent. Moreover, we call Model (\ref{non_causal_AR}), together with $\mathcal G(\mathcal V,\mathcal E)$, as non-causal graphical model. It is worth noting that such model is the symmetric version of the non-causal directed network model defined in \cite{M2012-WF}.

In what follows we want to face the following problem.
Consider the non-causal graphical model defined in (\ref{non_causal_AR}) where $\mathcal E$ and $n$ are known.
Given a finite length realization of $y(t)$ generated by (\ref{non_causal_AR}), say
$y^N:=\{y(1),\ldots,y(N)\}$,
we want to estimate the model parameters $H_k$ with $k=0,\ldots,n$.

Our idea is to tackle such problem by means of a covariance extension problem. The latter is defined as follows:
given the covariance lags sequence
\begin{eqnarray}
R_k=\mathbb{E}[y(t+k)y(t)^{\T}],\ k=0,1,\ldots,n, \non
\end{eqnarray}
or a selection of entries of them,
find an infinite extension
$R_{n+1},R_{n+2},\ldots$ such that, the series expansion
\begin{eqnarray}
\Phi(e^{{\rm{j}} \theta}):=\sum_{k=-\infty}^{\infty} R_ke^{-{\rm{j}} \theta k}\non
\end{eqnarray}
with $R_{-k}=R_k^{\T}$, is a spectral density. That is, we want to find $\Phi$ that matches the given covariance lags  sequence. Clearly, among the spectral densities matching the covariance lags sequence we want to choose a particular solution that is the one which takes the rational form in (\ref{Phi1}).
If we are able to solve the aforementioned problem, then an estimate of the spectral density (and thus of $H$) from $y^N$ can be obtained as follows:
we can find an estimate of $R_0,\ldots,R_n$ from the data $y^N$ as follows
\begin{eqnarray}
\label{hatR_k}
\hat{R}_k=\frac{1}{N}\sum_{t=1}^{N-k}{y}(t+k){y}(t)^{\T},\ k=0,\ldots,n.
\end{eqnarray}
Then, an estimate of the spectral density is the particular solution of the  covariance extension problem in which we replace $R_k$, with $k=0,\ldots,n$, with their estimates in (\ref{hatR_k}).

\section{Covariance extension for non-causal graphical models}
\label{sec:TS}
We first derive the covariance extension problem for non-causal graphical models and the particular solution we are looking for. In order to do that, we draw inspiration from the maximum entropy (ME) solution to the covariance extension problem for graphical models encoding conditional dependence relations introduced in \cite{6365751}.
We can attach to the process a graph $\mathcal G(\mathcal V,\mathcal{\bar{E}})$, where the vertex set $\mathcal V=\{1,\ldots,m\}$  represents
the components of the process and the edge set $\mathcal{\bar{E}}\subseteq \mathcal V\times \mathcal V$ describes the conditional dependence relations between the
process components.
Notice that, we have that $(i,i)\in \mathcal{\bar{E}}$ for $i=1,\ldots,m$.
Therefore, $\bar{\mathcal E}$ is different from $\mathcal E$, which corresponds to the non-causal graphical model, because the hypothesis $[H_k]_{ll}=0$, with $l=1\ldots m$ and $k=0\ldots n$, implies $(i,i)\notin \mathcal E$.
 The ME solution to the covariance extension problem for such graphical models is given as follows:
\begin{eqnarray}
\begin{split}
\label{ME}
\mathop{\rm{max}}\limits_{\Phi\in\mathbb S_m}\ &  \int {\rm{\log\det}}\Phi,
\\
{\rm{s.t.}}\  &\left[\int{\Phi}e^{k{\rm{j}}\theta}\right]_{li}=[{R}_k]_{li},\ k=0,1,\ldots,n,\\
&(l,i)\in \mathcal{\bar{E}},
\end{split}
\end{eqnarray}
where $\mathbb S_m$ denotes the family of bounded, coercive, $\mathbb C^{m\times m}$-valued spectral density functions on the unit circle,
and the shorthand notation
$\int f$ denotes the integration of the function $f$ taking place on the unit circle with respect to the normalized Lebesgue measure.
Problem (\ref{ME}) admits a unique solution which is the  power spectral density corresponding to an AR graphical model whose topology corresponds to $\mathcal{\bar{E}}$.

It is interesting to note that Problem (\ref{ME}) can be written as a minimum relative entropy covariance extension problem.
 Consider two $m$-dimensional jointly Gaussian stationary stochastic processes $x:=\{x(t), \; t\in\mathbb Z \}$ and $w:=\{w(t),\; t\in\mathbb Z\}$ whose mean is
equal to zero.
Let $\Phi_x$ and $\Phi_w$ denote the  power spectral density of $x$ and $w$, respectively.
The relative entropy rate, \cite{FMP-12}, between $\Phi_x$ and $\Phi_w$ is defined as
\begin{eqnarray}
D_{REL}(\Phi_x\|\Phi_w)&:=&
\frac{1}{2}\Big\{\int
\Big[{\rm{\log\det}}(\Phi_x^{-1}\Phi_w)\non\\
&&\ \ +{\rm{tr}}(\Phi_x\Phi_w^{-\T})\Big]-m\Big\}.\non
\end{eqnarray}
Taking $\Phi_x=\Phi$ and $\Phi_w=I_m$, we have
\begin{eqnarray}
D_{REL}(\Phi\|I_m)=\frac{1}{2}\Big\{\int\Big({\rm{\log\det}}\Phi^{-1}+{\rm{tr}}\Phi\Big)-m\Big\}.\non
\end{eqnarray}
 Moreover, since $(i,i)\in \mathcal{\bar{E}}$, the term $\int {\rm{tr}}\Phi={\rm{tr}}R_0=\sum_{i=1}^m[R_0]_{ii}$ is fixed.
Thus, the ME problem (\ref{ME}) is equivalent to the following problem:
\begin{eqnarray}
\begin{split}
\label{D_REL}
\mathop{\rm{min}}\limits_{\Phi\in\mathbb S_m}\ & D_{REL}(\Phi\|I_m),
\\
{\rm{s.t.}}\  &\left[\int{\Phi}e^{k{\rm{j}}\theta}\right]_{li}=[{R}_k]_{li},\ k=0,1,\ldots,n,\\
&(l,i)\in \mathcal{\bar{E}}.
\end{split}
\end{eqnarray}
It is important to point out that there is a difference between these two problems.
In Problem (\ref{ME}), if we replace $\mathcal{\bar{E}}$ with $\mathcal E$, where we recall that $(i,i)\notin \mathcal E$, then it is not difficult to see that Problem (\ref{ME}) does not admit solution.
 However, Problem (\ref{D_REL}) is still well-defined,
 that is a unique solution can be found even in the case we replace $\mathcal{\bar{E}}$ with $\mathcal E$, see \cite{LINKPRED}.
 In order to characterize the covariance extension problem we are looking for,
 we consider Problem (\ref{D_REL}) where $\mathcal{\bar{E}}$ is replaced with $\mathcal E$. Moreover, we have to substitute the objective function with another divergence index between $\Phi$ and $I_m$ which selects a solution of the form in (\ref{Phi1}).
 More precisely,
 we consider the transportation distance which is defined as follows.
\begin{Definition}
\label{De_tran}
Consider two $m$-dimensional stationary stochastic processes $x:=\{x(t), \; t\in\mathbb Z \}$ and $w:=\{w(t),\; t\in\mathbb Z\}$.
The latter are completely described by the finite dimensional
probability density functions $p_x(x_t, x_s; t, s)$ and $p_w(w_t, w_s; t, s)$ with $t,s \in \mathbb{Z}$.
Let $p_{x,w}(x_t, x_s, w_u, w_v; t,s,u,v)$ with
$t, s, u, v \in \mathbb{Z}$, be
the finite dimensional joint probability density of $x(t)$ and $w(t)$. The transportation cost (i.e. distance) is the variance of the discrepancy process $x(t)-w(t)$:
\begin{eqnarray}
\label{tr_cost}
d(p_x,p_w)=&\bigg[\mathop{\rm{inf}}\limits_{p_{x,w}\in \mathcal{P}}\mathbb{E}[\|x(t)-w(t)\|^2]\ \bigg]^{1/2},
\end{eqnarray}
where $\mathcal{P}$ denotes the set of Gaussian joint probability densities $p_{x,w}$ with marginals $p_{x}$ and $p_{w}$.
\end{Definition}
In \cite{Z-TRANSP} it was shown that
in the case the processes are zero mean and jointly Gaussian, the transportation distance between $x(t)$ and $w(t)$ defined in (\ref{tr_cost}) becomes
\begin{eqnarray}
\label{Hell_di}
\begin{split}
D_H(\Phi_x,\Phi_w)=\mathop{\rm{min}}\limits_{W_x\in\mathbb L_\infty^m}\ & {\rm{tr}}\int(W_x-W_w)(W_x-W_w)^{*},\\
{\rm{s.t.}}\  & W_xW_x^{*}=\Phi_x,
\end{split}
\end{eqnarray}
where:
$\mathbb L_\infty^m$ is the space of $\mathbb C^{m\times m}$-valued functions on the unit circle which are bounded almost everywhere; $\Phi_x$ and $\Phi_w$ are the power spectral densities of $x(t)$ and $w(t)$, respectively and $\Phi_w=W_wW_w^{*}$ where the superscript ${*}$ denotes the conjugate transpose.
Notice that $D_H(\Phi_x,\Phi_w)$ is also the Hellinger distance defined for spectral densities, \cite{FPR-08}.
Substituting the objective function and $\bar{\mathcal E}$ with the transportation distance and $\mathcal E$, respectively,
we obtain the following covariance extension problem
\begin{eqnarray}
\begin{split}
\label{tr_di_op1}
\mathop{\rm{min}}\limits_{\Phi\in\mathbb S_m}\ & D_H(\Phi,I_m),
\\
{\rm{s.t.}}\  &\left[\int{\Phi}e^{k{\rm{j}}\theta}\right]_{li}=[{R}_k]_{li},\ k=0,1,\ldots,n,\\
&(l,i)\in \mathcal E.
\end{split}
\end{eqnarray}
In the next section, we show that the solution to problem (\ref{tr_di_op1}) does exist and it corresponds to a symmetric non-causal graphical model with power spectral density of the form (\ref{Phi1}).

It is worth noting that $$\mathbb H_T(\Phi):=-D_H(\Phi,I_m)$$
can be understood as the entropy of the process with spectral density $\Phi$ induced by the transportation distance. Indeed,  an entropic index, like $\mathbb H_T(\Phi)$,  should measure how close the process with spectral density $\Phi$ is to normalized white noise with independent components, i.e. a  process having a spectral density  equal to $I_m$, see also
\cite{Zhu-Zorzi2023cepstral,TAC_ZHU_ZORZI}.

\section{Dual problem}
\label{sec:TS_dual}
We characterize the solution to Problem (\ref{tr_di_op1}) by means of the Lagrange multipliers theory. In view of (\ref{Hell_di}),
we take the spectral factor of $\Phi_w=I_m$ as $W_w=I_m$ and
the spectral factor of $\Phi_x=\Phi$ as $W_x=W$.
Then, problem (\ref{tr_di_op1}) can be rewritten as
\begin{eqnarray}
\begin{split}
\label{tr_di_op2}
\mathop{\rm{min}}\limits_{W\in\mathbb L_\infty^m}\ & {\rm{tr}}\int(W-I_m)(W-I_m)^{*},
\non\\
{\rm{s.t.}}\  &\left[\int WW^{*}e^{k{\rm{j}}\theta}\right]_{li}=[{R}_k]_{li}, \ k=0,1,\ldots,n,\non\\
 &(l,i)\in \mathcal E.\non
\end{split}
\end{eqnarray}
It is possible to prove that
the dual function (with opposite sign) is
\begin{eqnarray}
{J}(\mathbf H)
&=&{\rm{tr}}\int[(I_m-H)^{-1}-I_m]-\sum_{k=0}^n{\rm{tr}}(H_k^{\T}{R}_k),\non
\end{eqnarray}
where $\mathbf  H=[H_0\ H_1\ldots H_n]\in \mathcal{C}$ is the Lagrange multiplier such that  $H_k\in\mathbb{R}^{m \times m}$, $H_0=H_0^{\T}$, $[H_k]_{li}=0$ and $[H_k]_{il}=0$ for $(l,i)\notin \mathcal E$,
and the corresponding optimal form for the spectral density is
\begin{eqnarray}
\label{Phi2}
\Phi
=(I_m-H)^{-2},
\end{eqnarray}
which corresponds to the spectral density in (\ref{Phi1}).
Therefore, the dual problem is
\begin{eqnarray}
\label{tr_di_op}
\begin{split}
\mathop{\rm{min}}\limits_{\mathbf  H\in \mathcal{C}}\ &J(\mathbf  H),\\
{\rm{s.t.}}\ &I_m-H>0,\\
& [H]_{li}=0,\ (l,i)\notin \mathcal E.
\end{split}
\end{eqnarray}
If the dual problem admits a unique solution in the interior of the feasible set, say $\hat{\mathbf  H}$,
then the unique solution of the primal problem is $\hat{\Phi}
=(I_m-\hat{H})^{-2}$.

We conclude that the solution to Problem (\ref{tr_di_op1}) corresponds to a double-sided AR non-causal graphical model with topology $\mathcal E$.

\section{The ARMA case}
\label{sec:TS_ARMA}
A generalization of (\ref{non_causal_AR}) is the following ARMA model in which the AR part is double-sided, while the MA part is one-sided:
\begin{eqnarray}
\label{non_causal_ARMA}
y(t)&=&A(z)[I_m-H(z)]^{-1}e(t),
\end{eqnarray}
where $A(z):={\rm{diag}}(a_1(z),a_2(z),...,a_m(z))$ is a diagonal polynomial matrix  with elements in the main diagonal
\begin{eqnarray}
 a_l(z)=1+\sum_{k=1}^{p}a_{l,k}z^{-k}.\non
\end{eqnarray}
We assume that the roots of $a_k(z)$ for $k=1,\ldots,m$ are  inside the unit circle. From (\ref{non_causal_ARMA}), we have
\begin{eqnarray}
y(t)&=&\{I_m-[I_m-H(z)]A^{-1}(z)\}y(t)+e(t),\non\\
    &=&[I_m-A^{-1}(z)]y(t)+H(z)A^{-1}(z)y(t)+e(t).\non
\end{eqnarray}
Therefore the $l$-th component of $y(t)$ can be written as
\begin{eqnarray}
\label{y_l}
{y}_l(t)=\frac{a_l(z)-1}{a_l(z)}{y}_l(t)-\sum_{i\neq l}\frac{1}{a_i(z)}[H(z)]_{li}{y}_i(t)+e_l(t),\non
\end{eqnarray}
from which we can see that
the first term on the right hand side is a linear function of the past of itself, while the second one is a linear function of the other components of $y$. Also in this case we can attach to Model (\ref{non_causal_ARMA}) a symmetric directed graph $\mathcal G(\mathcal V,\mathcal E)$ with vertex set $\mathcal V=\{1,\ldots,m\}$  and edge set
$\mathcal E\subseteq \mathcal V\times \mathcal V$ reflecting the support of $H(z)$.
If we define $$ \mathcal Y_{\mathcal I,l,t}:=\overline{\mathrm{span}}\{ y_i(s) \hbox{ s.t. } \; (i,s)\in\mathbb Z_{l,t}\cup \mathbb Z_{\mathcal I}\}$$
with
\begin{align*}
\mathbb Z_{l,t}&:=\{\, (l,s) \hbox{ s.t. } s<t\}, \\
\mathbb Z_{\mathcal I}&:=\{\, (i,s) \hbox{ s.t. } s\in\mathbb Z, \, i\in \mathcal I\subset \mathcal V\},
\end{align*}
then $$ \hat y_l(t)=\mathbb E[ y_l(t)| \mathcal Y_{\mathcal V,l,t}],$$
which represents the estimate of $y_l(t)$ given the other components of $y$ and the past of $y_l$. Moreover,
$$ \hat y_l(t)=\mathbb E[ y_l(t)| \mathcal Y_{\mathcal V,l,t}]=\mathbb E[ y_l(t)| \mathcal Y_{\mathcal V\setminus\{i\},l,t}]$$
if and only if $[H(z)]_{li}\equiv 0$.

In what follows we want to learn from the collected data $y^N$
 the ARMA model (\ref{non_causal_ARMA}) where the orders $p,n$ and its associated graph $\mathcal G(\mathcal V,\mathcal E)$ are known.
More precisely, we want to estimate $H(z)$ and $A(z)$.

From (\ref{non_causal_ARMA}),
 we can define process $\xi:=\{\xi(t),\; t\in\mathbb Z\}$ as
 \begin{eqnarray}
 \label{non_causal_trans}
\xi(t):=A^{-1}(z)y(t)=[I_m-H(z)]^{-1}e(t).
\end{eqnarray}
Process $\xi$ is a double-sided AR process having the same structure of the one in (\ref{non_causal_AR}).
Accordingly, the idea is to split the optimization of $A(z)$ and $H(z)$ in two sequential steps as described below.\\
\textbf{Optimization of A}.
Notice that the power spectral density of the process ${y}_l:=\{{y}_l(t),\; t\in\mathbb Z\}$ is
$$\frac{|a_l(e^{{\rm{j}}\theta})|^2}{[(I_m-H(e^{{\rm{j}}\theta}))^{-2}]^{-1}_{ll}}.$$
Thus, to estimate $A(z)$ we consider a simplified model in which each component of $y$ is modeled independently.
More precisely,
we model the $l$-th component of $y$ as
\begin{eqnarray}
\label{si_model}
c_l(z){y}_l(t)=a_l(z)e_l(t),
\end{eqnarray}
where $c_l(z)=1+\sum_{k=1}^{n}c_{l,k}z^{-k}$,
and its power spectral density is
\begin{eqnarray}
\Phi_l(e^{{\rm{j}}\theta})&=&\frac{|a_l(e^{{\rm{j}}\theta})|^2}{|c_l(e^{{\rm{j}}\theta})|^2}.\non
\end{eqnarray}
In plain words, we approximate $[(I_m-H(z))^{-2}]^{-1}_{ll}$ with $|c_l(z)|^2$.
Thus, for $l=1,\ldots,m$, we estimate $a_l(z)$ through Model (\ref{si_model}) using $y_l^N:=\{y_l(1)\ldots y_l(N)\}$  by means of the prediction error method (PEM), \cite{SI99}.

\textbf{Optimization of H}. In regard to $H(z)$,
we can obtain its estimate using the data set $y^N$ and $A(z)$ (the latter has been estimated in the previous step).
More precisely, consider
process $\xi$ defined in (\ref{non_causal_trans}). Then, we can compute a finite
length trajectory of it, say $\xi^N$, by passing $y^N$ through the filter
$A^{-1}(z)$.
Since $\xi$ is a double-sided AR process with spectral density of the form in (\ref{Phi1}), we can find an estimate of $H(z)$ as the solution  of the dual problem in (\ref{tr_di_op}) where the covariance lags are the ones of $\xi$.

The resulting identification procedure is given in Algorithm \ref{Algorithm}.
\begin{algorithm}[!htb]
\caption{Estimator for the ARMA case}
\label{Algorithm}
{\bf Input:} $y^N$, $p$, $n$ and $\mathcal E$

{\bf Output:} $H(z)$ and $A(z)$
\begin{algorithmic}[1]
\smallskip
\STATE
Estimate $a_l(z)$, with $l=1,...,m$  using PEM.
\STATE ${A}(z)={\rm{diag}}({a}_1(z),{a}_2(z),\ldots,{a}_m(z))$.
\smallskip
\STATE Let $\xi(t):={A}(z)^{-1}y(t)$, compute $\xi^N$ using $y^N$.
\smallskip
\STATE
Compute $\hat{R}_k=\frac{1}{N}\sum_{t=1}^{N-k}{\xi}(t+k){\xi}(t)^{\T}$.
\smallskip
\STATE Compute $H(z)$ as the solution of (\ref{tr_di_op}). \end{algorithmic}
\end{algorithm}

\section{Numerical experiments}
\label{sec:Simulation}
To provide empirical evidence of the estimation performance
of the algorithm, simulations studies have been
performed by using the software Matlab-R2020b.

\subsection{AR case}
\label{sec:Simulation1}
We consider a Monte Carlo experiment which is structured
as follows. We generate 100 double-sided AR models (\ref{non_causal_AR}) of dimension $m=15$,
of order $n=2$, whose fraction of nonnull entries in $H$ is equal to 0.1.
The position of such nonnull entries is chosen randomly for each model. For each model we generate
a finite data sequence of length $N$ and
consider the following estimators:
\begin{itemize}
\item
TE is the estimator defined as the maximum transportation entropy solution of the covariance extension problem (\ref{tr_di_op1}).
\item TE-F
is the maximum transportation entropy estimator with full graph $\mathcal G(\mathcal V,\mathcal E_{f})$, where
 $\mathcal E_{f}= \mathcal V\times \mathcal V\backslash \{ (i,i),\ i\in \mathcal V\}$.
 Thus, TE-F is the estimator obtained by solving the covariance extension problem (\ref{tr_di_op1}) with $\mathcal E$ replaced by $\mathcal E_{f}$.
\item
ME is the maximum entropy estimator, see \cite{burg1975maximum}, where the order of the AR process is set equal to $n$.
\end{itemize}
For each estimator,
we compute the relative error in the estimation of the
AR parameters as $$e=\frac{\|\hat{\bfH}-\bfH\|}{\|\bfH\|},$$
where $\|\cdot\|$ denotes the $2$-norm, $\hat{\bfH}=[\hat{H_0}\ \hat{H_1}\cdots \hat{H_n}]$ is the estimate and $\bfH=[H_0\ H_1\cdots H_n]\in \mathbb{R}^{m(n+1)\times m}$ is the true value.

\begin{figure*}[!hbt]
    \centering
    \begin{subfigure}{0.32\linewidth}
    \centering
        \includegraphics[width=\hsize]{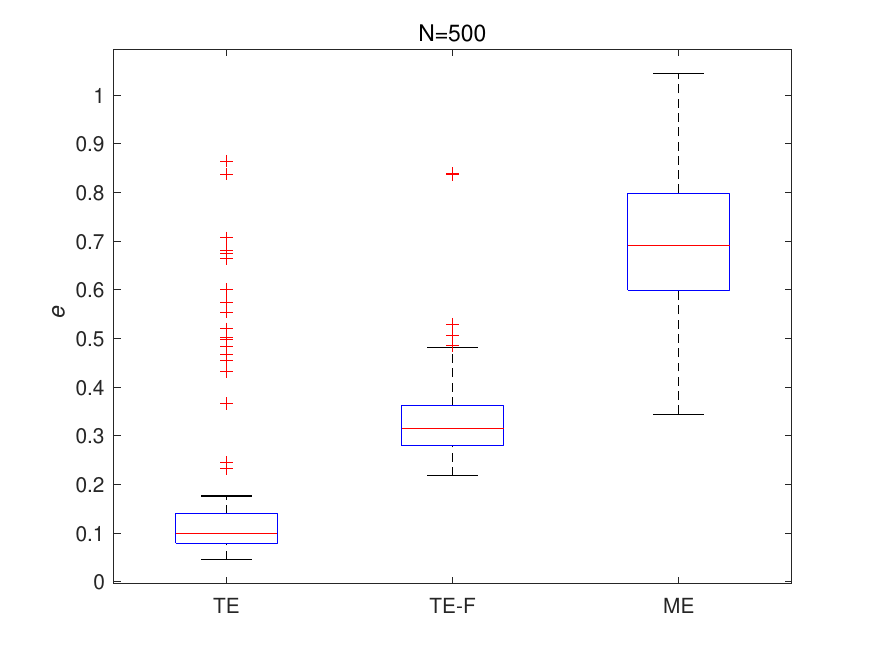}
        \label{Fig_non_causal_AR_N500}
    \end{subfigure}
   \centering
    \begin{subfigure}{0.32\linewidth}
    \centering
        \includegraphics[width=\hsize]{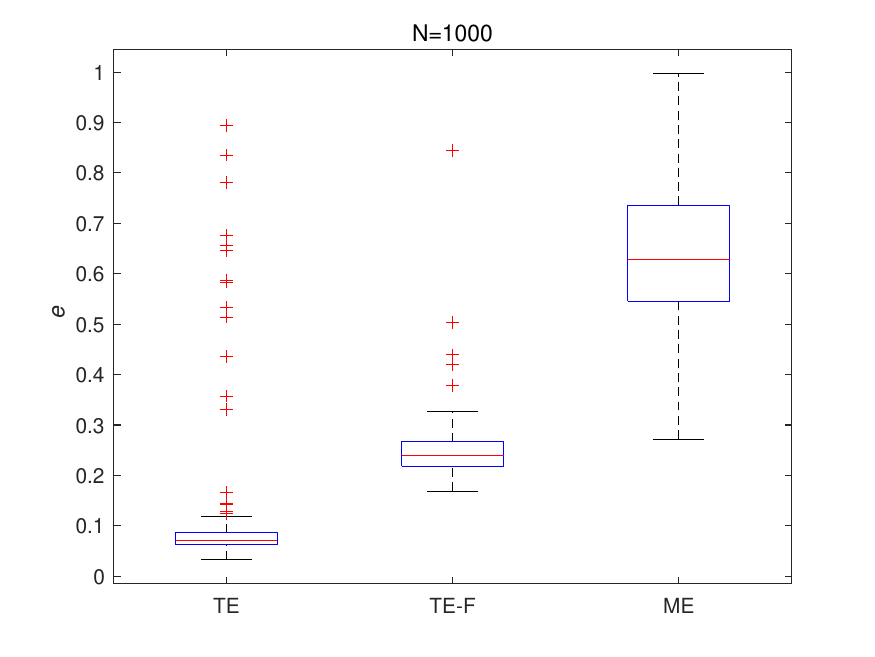}
        \label{Fig_non_causal_AR_N1000}
    \end{subfigure}
  \centering
    \begin{subfigure}{0.32\linewidth}
    \centering
        \includegraphics[width=\hsize]{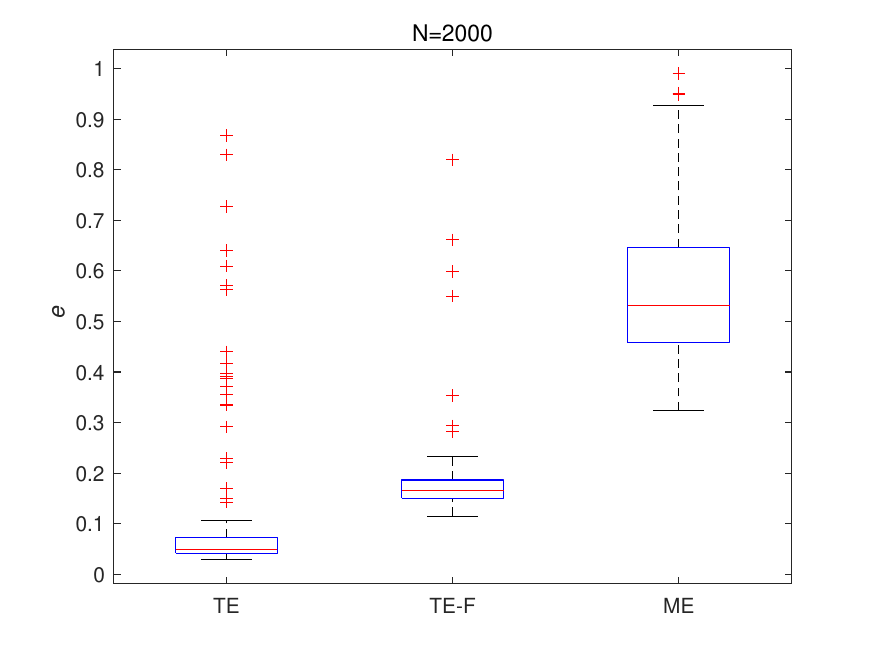}
        \label{Fig_non_causal_AR_N2000}
    \end{subfigure}
    \caption{Monte Carlo experiments in the AR case using different data lengths ($N=500$, $N=1000$ and $N=2000$).}
    \label{Fig_non_causal_AR}
\end{figure*}

We have considered three different lengths of the data, that is $N=500$, $N=1000$ and $N=2000$.
The boxplots of the relative error $e$ for different data lengths are depicted in Fig. \ref{Fig_non_causal_AR}.
For the case $N=500$ we see that TE exhibits the best performance thanks to the prior knowledge of the graph topology, allowing it to estimate fewer parameters;
ME exhibits the worst performance because it searches the best model in a model  class which is different from the one  corresponding to (\ref{non_causal_AR}).
For the cases $N=1000$ and $N=2000$, the scenario does not change.
Moreover, as expected, the estimation error decreases as the data length increases for all the estimators.

\subsection{ARMA case}
\label{sec:Simulation2}
 We consider a Monte Carlo experiment in which 100 double-sided ARMA models are generated by (\ref{non_causal_ARMA}) with the same dimension and AR order as in the AR case in Section \ref{sec:Simulation1} and with the MA order $p=1$.
The fraction and position selection of nonnull entries in $H$ are also consistent with the settings described in Section \ref{sec:Simulation1}.

\begin{figure*}[!hbt]
    \centering
    \begin{subfigure}{0.32\linewidth}
    \centering
        \includegraphics[width=\hsize]{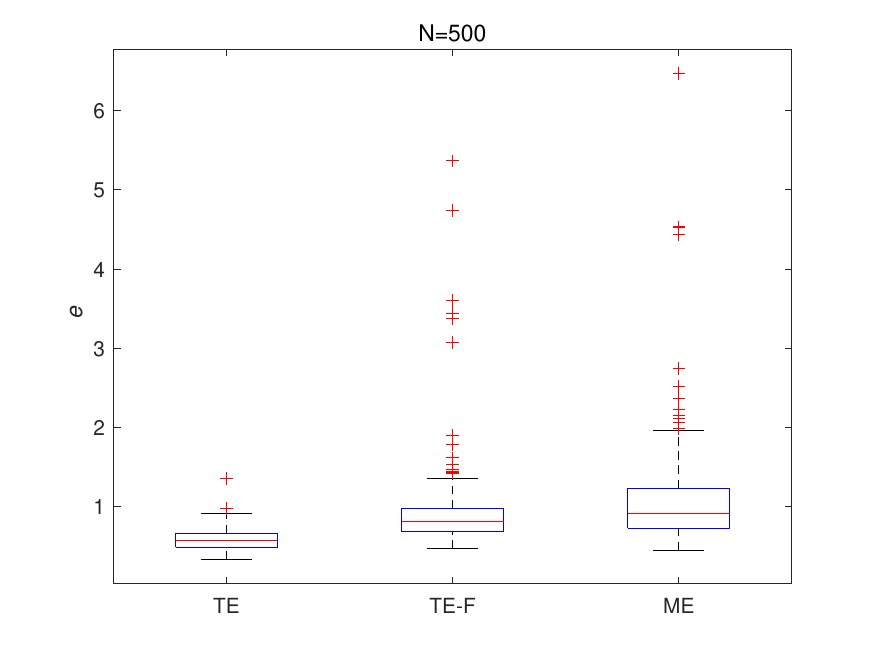}
        \label{Fig_non_causal_ARMA_N500}
    \end{subfigure}
   \centering
    \begin{subfigure}{0.32\linewidth}
    \centering
        \includegraphics[width=\hsize]{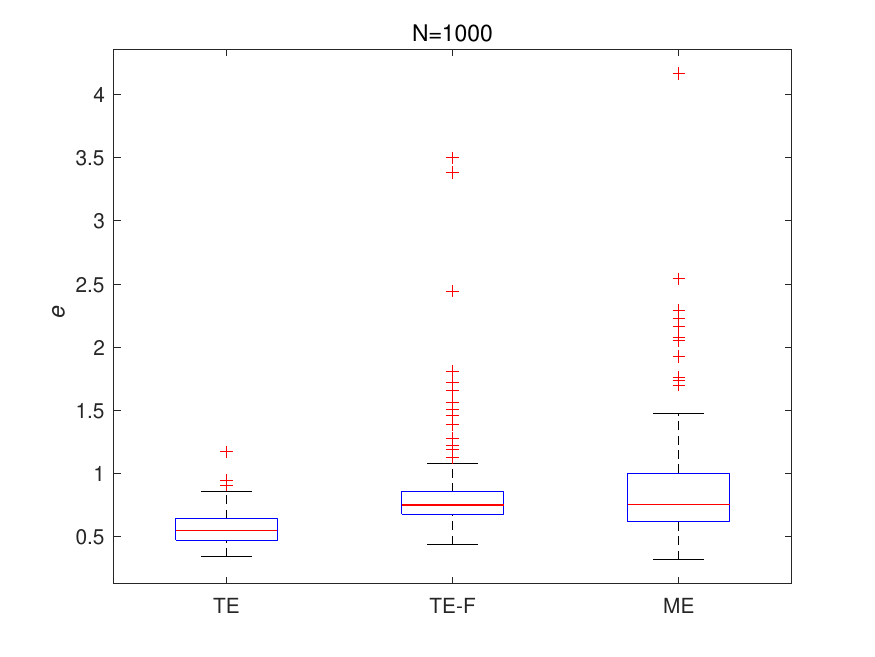}
        \label{Fig_non_causal_ARMA_N1000}
    \end{subfigure}
  \centering
    \begin{subfigure}{0.32\linewidth}
    \centering
        \includegraphics[width=\hsize]{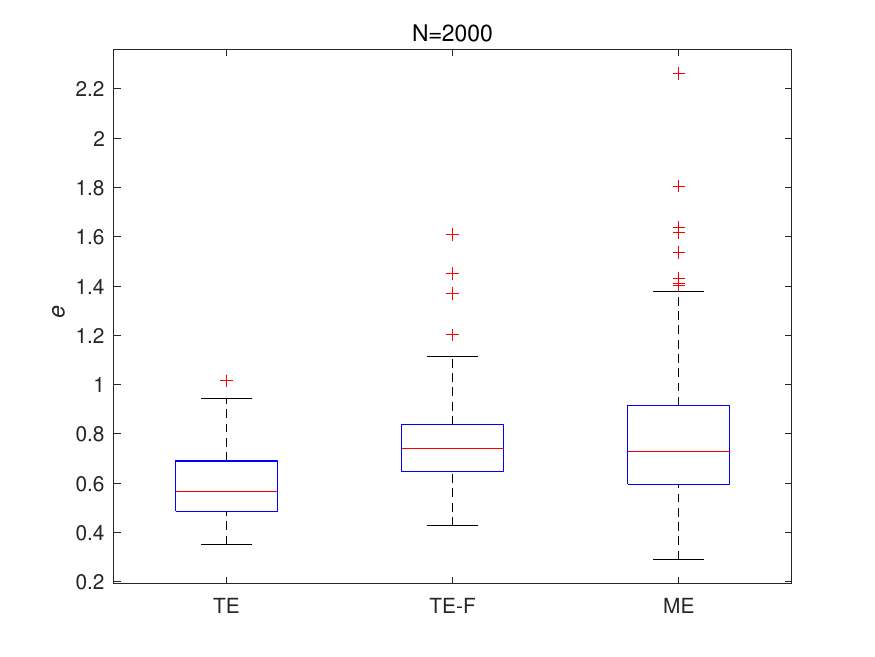}
        \label{Fig_non_causal_ARMA_N2000}
    \end{subfigure}
    \caption{Monte Carlo experiments in the ARMA case using different data lengths ($N=500$, $N=1000$ and $N=2000$).}
    \label{Fig_non_causal_ARMA}
\end{figure*}

We consider the following  estimators in order to identify the model:
\begin{itemize}
\item  TE is the estimator presented in Section \ref{sec:TS_ARMA}, see Algorithm \ref{Algorithm}.
 \item TE-F is the estimator presented in Section \ref{sec:TS_ARMA} where $\mathcal E$ replaced by $\mathcal E_{f}$, i.e. we estimate a model corresponding to a full graph.
 \item ME is the maximum entropy version of Algorithm \ref{Algorithm}, i.e.  step 5 is replaced with the maximum entropy estimator \cite{burg1975maximum}.
 \end{itemize}
  As in Section \ref{sec:Simulation1}, we have considered three different data lengths $N=500$, $N=1000$ and $N=2000$.
 The boxplots of the relative error $e$ for these three cases are depicted in Fig. \ref{Fig_non_causal_ARMA}.
As we can see, the estimation errors of TE-F and ME are comparable,
while TE outperforms TE-F and ME because it exploits the a priori information regarding the graph topology.

\section{Conclusions}
\label{sec:Conclusions}
In this paper we have considered the problem to identify
symmetric non-causal graphical models.
More precisely, we have characterized the maximum transportation entropy solution to a covariance extension problem
and have shown that this solution corresponds to a double-sided AR non-causal graphical model.
We have also extended this idea to a class of ARMA models.
Finally we have performed some numerical experiments showing the effectiveness of the proposed method.
Notice that
in real situations, the graph topology is unknown.
Therefore, these results can be regarded as the first step towards a regularized identification paradigm in order to learn from the data also the graph topology.
In plain words,
in the future we aim to generalize the proposed paradigm which jointly estimates the graphical structure and the parameters.

\bibliographystyle{ieeetr}
\bibliography{references}

\end{document}